\documentclass[conference]{IEEEtran}
\IEEEoverridecommandlockouts
\usepackage{cite}
\usepackage{amsmath,amssymb,amsfonts}
\usepackage{algorithmic}
\usepackage{graphicx}
\usepackage{textcomp}
\usepackage{xcolor}
\usepackage{subfigure}
\usepackage{color}

\newcommand{\ie}{\textit{i.e.,} }
\newcommand{\eg}{\textit{e.g.,} }

\def\BibTeX{{\rm B\kern-.05em{\sc i\kern-.025em b}\kern-.08em
    T\kern-.1667em\lower.7ex\hbox{E}\kern-.125emX}}
\begin{document}

\title{Edge Sparse Basis Network: A Deep Learning Framework for EEG Source Localization \\

\thanks{This research was supported by National Natural Science Foundation of China (No. 62001205), Guangdong Natural Science Foundation Joint Fund (No. 2019A1515111038), Shenzhen Key Laboratory of Smart Healthcare Engineering (ZDSYS20200811144003009)}
\thanks{Chen Wei and Kexin Lou are the co-first authors;\\  $^*$ Quanying Liu is the corresponding author.}
}

\author{\IEEEauthorblockN{Chen Wei $^{\dagger}$; Kexin Lou $^{\dagger}$}
\IEEEauthorblockA{\textit{Shenzhen Key Laboratory of Smart Healthcare Engineering}\\
\textit{Department of Biomedical Engineering} \\
\textit{Southern University of Science and Technology}\\
Shenzhen, China\\
\{weic3;12063004\}@mail.sustech.edu.cn}

\and
\IEEEauthorblockN{Zhengyang Wang}
\IEEEauthorblockA{\textit{Department of Neurology \& Anatomy} \\
\textit{Wake Forest School of Medicine, Winston-Salem}\\
North Carolina, United States\\
zhewang@mail.wakehealth.edu}

\and
\IEEEauthorblockN{Mingqi Zhao; Dante Mantini}
\IEEEauthorblockA{\textit{Movement Control and Neuroplasticity Research Group}\\
\textit{KU Leuven}\\
Leuven, Belgium\\
\{mingqi.zhao; dante.mantini\}@kuleuven.be}
\and
\IEEEauthorblockN{Quanying Liu $^*$}
\IEEEauthorblockA{\textit{Shenzhen Key Laboratory of Smart Healthcare Engineering}\\
\textit{Department of Biomedical Engineering} \\
\textit{Southern University of Science and Technology}\\
Shenzhen, China\\
liuqy@sustech.edu.cn}
}

\maketitle

\begin{abstract}

EEG source localization is an important technical issue in EEG analysis. Despite many numerical methods existed for EEG source localization, they all rely on strong priors and the deep sources are intractable.
Here we propose a deep learning framework using spatial basis function decomposition for EEG source localization. This framework combines the edge sparsity prior and Gaussian source basis, called Edge Sparse Basis Network (ESBN). The performance of ESBN is validated by both synthetic data and real EEG data during motor tasks. The results suggest that the supervised ESBN outperforms the traditional numerical methods in synthetic data and the unsupervised fine-tuning provides more focal and accurate localizations in real data. Our proposed deep learning framework can be extended to account for other source priors, and the real-time property of ESBN can facilitate the applications of EEG in brain-computer interfaces and clinics.

\end{abstract}


\section{Introduction}
\label{S:introduction}

Electroencephalography (EEG), as a noninvasive neural signal acquisition technique, can record electrical potential signals from human scalp. With the characteristics of low cost, strong portability, and high time resolution, it has been widely used spanning from fundamental research in cognitive neuroscience to engineering applications in brain computer interfaces (BCI). EEG also has a high value in clinical applications of neurological diseases, such as depression and epilepsy. However, since the current EEG Source Imaging (ESI) is mostly based on low-channel EEG signals (such as 32 or 64 channels of EEG) and thus the number of brain sources is much higher than the number of electrodes, EEG source tracing is therefore an ill-posed problem. It is difficult to solve EEG source localization problem effectively. 

EEG source localization involves two problems: 1) the EEG forward problem, which is to build a head volume conductivity model for describing how the electrical signals of the brain signal source are transmitted to the scalp electrodes~\cite{liu2018detecting}; 2) the EEG inverse problem, that is, to estimate the most possible source activity which could generate the scalp EEG signals~\cite{grech2008review}. Solving the forward problem is usually the prerequisite of the inverse problem.

To solve the EEG forward problem, it first needs to construct a head model, that is, to establish a conductivity model through the structure of the brain source in the human head (solution space) and electrical conduction (forward operator)~\cite{RN935}. Then several numerical algorithms can be used to obtain the solution of the EEG forward problem, mainly including the boundary element method (BEM), finite element method (FEM) or finite difference method (FDM).

The EEG inverse problem is to estimate the intensity and distribution of neural activity sources based on the forward solution~\cite{acar2013effects}. However, since the number of neural sources is far more than the number of electrodes, the solution of the inverse EEG problem is ill-posed, which means it has many possible solutions. To constrain the solution space, the numerical methods for EEG inverse problem have to add prior information or regularization~\cite{RN934}. Consequently, they rely heavily on the formulation of the regularization terms, which reflect simplified prior knowledge of brain sources. However, the real EEG sources might not be simply formulated in mathematical terms, such as the L1-norm~\cite{uutela1999visualization} or L2-norm~\cite{RN927}. Also, some numerical methods that cooperated with multiple regularizations are hard to express explicitly and are time-consuming ~\cite{abeyratne1991artificial}, which severely limits the power of EEG on the real-time BCI. 

Here we propose Edge Sparse Basis Network (ESBN) to solve the EEG inverse problem in a deep learning and data-driven manner. ESBN combines edge sparse priors with the spatial basis functions derived from the data, which allows to reconstruct the EEG source dynamics in real-time. Although ESBN is trained on synthetic data, we verify it with both simulated data and real EEG dataset.
The three main contributions of this paper are summarized as following.
\begin{itemize}
  \item The framework of ESBN allows for bidirectional flow of information: 1) generate EEG data from EEG source and 2) reconstruct EEG source from EEG data.
  \item ESBN has an end-to-end supervised version based on synthetic data and an unsupervised version to fine tune the model based on real EEG data for better generalization.
  \item The supervised learning ESBN achieves state-of-the-art (SOTA) performance in synthetic data, and the unsupervised ESBN offers better performance in real data.
\end{itemize}



\section{Related work}
\label{S:related_work}

\subsection{Numerical Solutions for EEG inverse problem}
There is a long history to study numerical algorithms for inverse problems, including EEG inverse problem. One of the most famous numerical solutions is minimal norm estimation (MNE). The goal of MNE is to minimize the difference between the estimated source current and real source current. However, due to the ill-posed nature of the EEG inverse problem, prior assumptions on mathematical, anatomical, and biophysical constraints are required to be incorporated into the basic MNE method~\cite{RN934,sohrabpour2016imaging}. Many numerical methods for EEG inverse problem have been proposed based on some simple assumptions, including the smoothness~\cite{pascual1994low}, spatial sparsity\cite{phillips2002anatomically,friston2008multiple,bore2018sparse}, edge sparsity~\cite{sohrabpour2016imaging,sohrabpour2020noninvasive}, as well as the time-frequency characteristics~\cite{gramfort2013time} of EEG sources. 


These Numerical methods are mainly divided into three categories: non-parametric methods, parametric methods~\cite{grech2008review} and Bayesian methods~\cite{RN941}. Non-parametric methods include MNE and its variants (\eg weighted MNE). Because the solution of these methods can be expressed as a linear operator, the calculation speed is much faster compared with the parametric methods. In contrast, Beamforming method, as a well-known parametric method, relies on multiple iterations and thus is computationally expensive. Consequently, it is not suitable for real-time applications. 

Although numerical methods of EEG inverse problem have been verified in many studies~\cite{liu2018detecting,mahjoory2017consistency,grech2008review,RN941}, they still have some limitations. First, the prior distribution of the actual brain sources is very complicated due to the irregularity of the head shape and brain tissue structure. Thus using a simple prior to express the source distribution in numerical methods is oversimplified. Second, the numerical solutions heavily rely on the data quality, and the performance drops dramatically with the noise. Third, the deep sources are hardly reconstructed due to the ill-posed nature and the regularizations added in numerical methods. 

\subsection{Deep learning methods}

Deep neural networks have been considered a potential tool for inverse problems. Theoretically, it has been proven that deep neural networks are able to fit any distribution. In practice, many network structures have been proposed, aiming to obtain multi-scale source information from the original data. In this way, the deep learning models can adapt to more complex distributions and thus have greater potential to generate more realistic source distributions compared with traditional numerical algorithms. 

Some pioneer studies have tried to bring shallow artificial neural networks into EEG inverse problem~\cite{van2000eeg,jun2002fast,jun2005fast}. However, these studies are limited by the sample size, network depth and computational power at that time, leading to poor performance. In recent years, with the rapid development of deep learning algorithms, great progress has been made in solving ill-posed inverse problems, such as remote sensing\cite{RN938}, physics\cite{RN939,RN940}, and medical imaging\cite{RN936,RN937}. For instance, convolutional neural network (CNN) has been used to locate the pacemaker of premature cardiac beats based on 12-lead ECG~\cite{bollmann2019deepqsm}. The U-Net on functional magnetic resonance images (fMRI) for locating the source of midbrain~\cite{graves2013speech}. However, so far only a few deep learning methods have been proposed for EEG inverse problem. Among them, Multi-Layer Perceptron(MLP) network~\cite{pantazis2020meg}, various CNN networks, such as UNet and Generative Adversarial Network (GAN) ~\cite{pantazis2020meg,hecker2020convdip,razorenova2020deep,sun2020sifnet} were used to solve stationary EEG inverse problem. Some studies ~\cite{dinh2019contextual,cui2019eeg} further combined long short term memory(LSTM) architecture to integrate temporal information. The potential of deep learning is largely underestimated in this field. We therefore focus on developing a novel deep learning framework for EEG source localization.

\section{Method}
\label{S:method}

\subsection{EEG inverse problem}

EEG forward problem can be mathematically described as
\begin{equation}
\begin{aligned}
\varPhi = K j +n
\end{aligned}
\label{eq:forward}
\end{equation}
where $j$ is the EEG source current with size $N \times T$; $\varPhi$ is the current density in the scalp EEG, with size $M \times T$; $K$ is the leadfield matrix with size $M \times N$; $n$ is the EEG channel noise; $M$ and $N$ are the number of EEG channels and sensors respectively. 

The inverse problem is to estimate the EEG source current $j$ based on the observed EEG data $\varPhi$. MNE and its variants estimate source currents $j^{MNE}$ based on covariance matrix of channel noise $C$ and the source-level covariance matrix $R$. Here $R$ is a Gaussian source distribution prior.

\begin{equation}
\begin{aligned}
j^{MNE} & = RK^{T}(KRK^{T}+\lambda^2C)^{-1}\varPhi \\
& = W\varPhi
\end{aligned}
\label{eq:MNE1}
\end{equation}
where $W = RK^{T}(KRK^{T}+\lambda^2C)^{-1}$ is the numerical inverse operator.
This solution can also be explicitly represented by:
\begin{equation}
\begin{aligned}
j^{M N E} =&\underset{j}{\operatorname{argmin}}\left\{\left\|(\phi-K j)\right\|_{C^{-1}}^{2}+\lambda^{2}\left\|j\right\|_{R^{-1}}^{2}\right\}
\end{aligned}
\label{eq:MNE2}
\end{equation}
Here, $\left\|P \right\|_{C^{-1}} = \sqrt{\text{tr}\left\{P^TC^{-1}P\right\}}$ denotes the Mahalanobis distance, and $\lambda$ is a regularization constant. ${\cal L}({\cal K} \left(j\right),\varPhi)$ here denotes the difference between the original EEG data and the projected EEG data from the estimated source using the forward model.
We can add regularization term $\cal S (j)$ into the objective function:
\begin{equation}
\begin{aligned}
\min_{j \in X}\left[{\cal L}({\cal K} \left(j\right),\varPhi)+\lambda{\cal S}(j)\right] &  &\text{for\  a\ fixed\ }\lambda \geq 0
\end{aligned}
\label{eq:loss-reg}
\end{equation}
which equals to reconstruct a mapping function: ${\cal K^{\dagger}_{\boldsymbol{\Theta}}} : Y\rightarrow X$ satisfying the pseudo-inverse property:
\begin{equation}
\begin{aligned}
{\cal K^{\dagger}_{\boldsymbol{\Theta}}}\left(\varPhi\right)\approx j_{true}
\end{aligned}
\label{eq:mapping}
\end{equation}
To be noted, the regularization term $\cal S (j)$ is not specified here. It can be defined based on the prior knowledge of EEG source.

In inverse problems, machine learning approaches parameterize the pseudo-inverse operators by a vector of parameters $ {\boldsymbol{\Theta}} \in Z$, and $ {\boldsymbol{\Theta}}$ can be learned using gradient descent during training\cite{adler2017solving}. 

For supervised learning when we have the ground-truth of EEG source $j$, the loss function can be represented by:
\begin{equation}
\begin{aligned}
{ L\left(\boldsymbol{\Theta}\right)}=\left\|\mathcal{K}_{\Theta}^{\dagger}(\Phi)-j\right\|
\end{aligned}
\label{eq:loss-supervised}
\end{equation}
where j is known if we use the synthetic data for training.

For the unsupervised learning applied to the real data when we do not know the ground-truth of EEG source, the loss function can be formulated as
\begin{equation}
\begin{aligned}
L\left(\boldsymbol{\Theta}\right)  ={\cal L}\left({\cal K} \left({\cal K^{\dagger}_{\boldsymbol{\Theta}}}\left(\varPhi\right)\right),\varPhi\right)+{\cal S}\left({\cal K^{\dagger}_{\boldsymbol{\Theta}}}\left(\varPhi\right)\right)
\end{aligned}
\label{eq:loss-unsupervised}
\end{equation}
where $\cal S$ is the regularization function that encodes prior information about $j_{true}$ and penalizes unfeasible solutions. Analogous to the loss for an autoencoder, $\cal L$ is the dissimilarity between the observed data and estimation in the sensor space.

\subsection{Edge sparse basis network}
\label{sub:edge}

We propose the Edge sparse basis network (ESBN) for real-time EEG source localization by bringing the concept of edge sparsity into the deep learning framework, as previous studies have shown that the edge sparsity regularization on EEG sources can improve EEG source localization performance~\cite{sohrabpour2016imaging,sohrabpour2020noninvasive}. ESBN is based on three presuppositions: 
\begin{enumerate}
\item The EEG activated sources can be represented by a linear combination of independent basis functions in the source space (as Eq.\ref{eq:basis_hypo}).
\item The brain state in each time can be expressed by a small set of basis functions. ($\cal M$ has a small rank.)
\item These basis functions have sparse edges with high gradients~\cite{sohrabpour2016imaging,sohrabpour2020noninvasive}. ($\Omega$ is edge sparse.)
\end{enumerate}

\begin{figure}[h]
\centering
\includegraphics[width=0.45\textwidth]{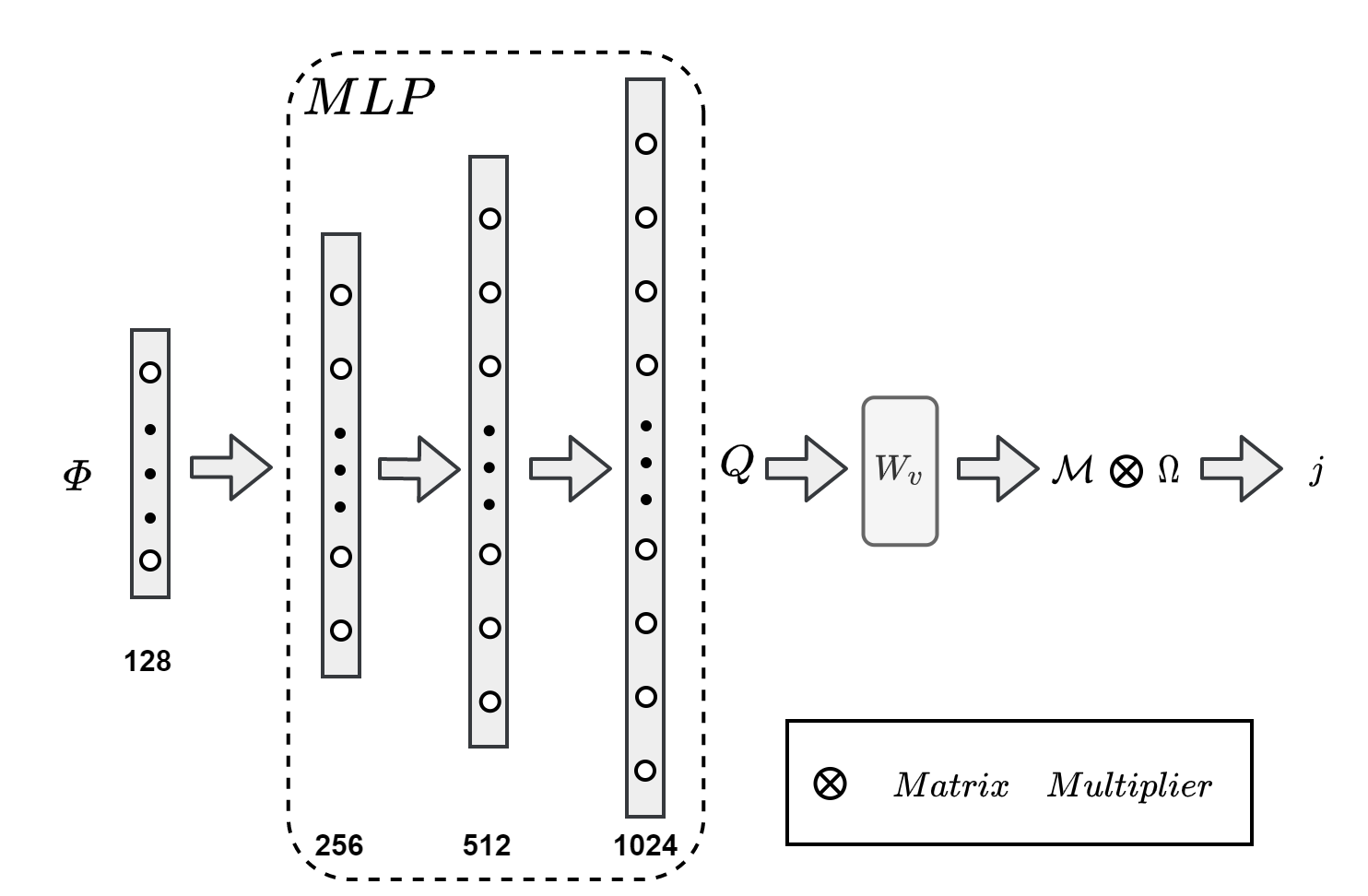}
\caption{Network structure for ESBN, the meaning of each term will be clarified in Section~\ref{sub:edge} }
\label{fig:Panel}
\end{figure}

These three presuppositions can be formulated as following:
\begin{equation}
\begin{aligned}
j_{true} ={\cal M} {\Omega} 
\end{aligned}
\label{eq:basis_hypo}
\end{equation}
where $\cal M$ is the weight of basis functions. $\Omega$ is the basis functions, which is edge sparse. $\Omega$ can learn from the training data using gradient descent:
\begin{equation}
\begin{aligned}
{\cal M} = F_{\boldsymbol{\Theta}}\left({\varPhi}\right)
\end{aligned}
\label{eq:Mu_hypo}
\end{equation}
where the notation $\Theta$ is the parameters of network $F$. As shown in Fig.~\ref{fig:Panel}, the specific calculation of $F$ is as follows:

\begin{equation}
\begin{aligned}
\cal{Q} &= MLP{\left({\varPhi}\right)} \\
\end{aligned}
\label{eq:Mu_attention}
\end{equation}
\begin{equation}
\begin{aligned}
\cal{M} &= W_{v}\cal{Q} \\
\end{aligned}
\label{eq:weightcoeffi_v}
\end{equation}

During the training, both linear and MultiLayer Perceptron (MLP) models work fine. In order to pursue better representation ability under the constrain of GPU memory, we choose a three layer MLP. Here ${\cal Q}$ is the feature extracted by $MLP$, $W_{{\cal V}}$ is the coefficient matrix for inferring weight matrix${\cal M}$. By multiplying the inferred weight of basis functions, we are able to estimate the EEG source $\hat{j}$: 
\begin{equation}
\begin{aligned}
\hat{j} &= F_{\boldsymbol{\Theta}}\left({\varPhi}\right)\Omega = {\cal K}^{\dagger}_{\boldsymbol{\Theta},\boldsymbol{\Omega}}\left(\varPhi\right)\\
\end{aligned}
\label{eq:basis_representation}
\end{equation}

The network can be represented by an inverse operator ${\cal K}^{\dagger}_{\boldsymbol{\Theta},}\left(\varPhi\right)$, with $ \boldsymbol{\Theta}$ and $\boldsymbol{\Omega}$ act as the targets of parameter optimization. In other words, the basis functions $\boldsymbol{\Omega}$ is also learnable.

The network is trained primarily on synthetic data (See  Eq.~\ref{eq:synthetic}) in an end-to-end fashion. The loss function for the supervised learning is defined as below: 
\begin{equation}
\begin{aligned}
{ L\left(\boldsymbol{\Theta}\right)}=\left\|F_{\boldsymbol{\Theta}}\left({\varPhi}\right)\Omega-j\right\|^{2}_{F} + {\cal {S}}\left({\cal M}, {\Omega} \right)
\end{aligned}
\label{eq:ESBN_SUPERVISED}
\end{equation}

Then the network is further trained with the real EEG data in an unsupervised way. In this way, the loss function is constructed at sensor level and fine tuned the basis functions with regularization terms on the weight of basis functions. By substitute Eq.\ref{eq:basis_representation} into Eq.\ref{eq:loss-unsupervised}, the loss function for the unsupervised learning can be represented as:
\begin{equation}
\begin{aligned}
L\left(\boldsymbol{\Theta}\right) = \left\|\varPhi-\mathcal{K}\mathcal{F}_{\boldsymbol{\Theta}} (\varPhi) \Omega \right\|^{2}_{C^{-1}} + {\cal {S}}\left({\cal M}, {\Omega} \right)
\end{aligned}
\label{eq:ESBN_UNSUPERVISED}
\end{equation}
Here $\mathcal{S}(\mathcal{M}, \Omega) = \mathcal{S}_{1}(\mathcal{M}) + \mathcal{S}_{2}(\Omega)$. We define $\mathcal{S}_{1}(\mathcal{M})$ , $\mathcal{S}_{2}(\Omega)$ and similarity as following: 
\begin{equation}
\begin{aligned}
\mathcal{S}_{1}(\mathcal{M})=\|{\cal M}\|_{1}
\end{aligned}
\label{eq:regularization on mu}
\end{equation}
\begin{equation}
\begin{aligned}
\mathcal{S}_{2}\left( {\Omega} \right) = \|V\Omega\|_{1}+\sum_{i,j}similarity(\Omega_{i},\Omega_{j})
\end{aligned}
\label{eq:regularization on gamma}
\end{equation}
\begin{equation}
\begin{aligned}
similarity(\Omega_{i},\Omega_{j}) = \frac{\Omega_{i}\Omega_{j}}{\|\Omega_{i}\|_{2}\|\Omega_{j}\|_{2}}
\end{aligned}
\label{eq:explain on similarity}
\end{equation}
where $\mathcal{S}_{1}\left(\cal  {M}\right)$ is to penalize the weight matrix of basis functions and constrains the number of activated basis; $\mathcal{S}_{2}(\Omega)$ is to penalize the basis functions, making them more sparse. The matrix $V$ in Eq.\ref{eq:regularization on gamma} is adapted from from~\cite{sohrabpour2016imaging}.

Since the number of edges in a volumetric head model is relatively large, leading to a high dimensional edge matrix. Therefore, we use three-dimensional Prewitt operator \cite{dhankhar2013review} as an edge extractor of 3D source images.

\section{Experiments}
\label{S:experiment}

To validate our method, we run experiments on both synthetic data and real EEG data. The synthetic data is simulated using Gaussian sparse sources~\cite{RN921} and 12-layer realistic head model~\cite{liu2018detecting}. The real EEG data is recorded during motor tasks~\cite{RN918}, including self-paced hand movement and foot movement.

\subsection{Experiments on synthetic data}
We generate a synthetic EEG data set, which has 450,000 EEG source and scalp EEG pairs in total, under different Signal Noise Ratio (SNR) conditions and dipole orientation settings. 
The EEG sources are sampled from the Gaussian sparse sources~\cite{RN921}. The source basis is defined as below:
\begin{equation}
\begin{aligned}
\mu_{n}(x)=\omega_{n}\left(\sqrt{2 \pi} \sigma_{s}\right)^{-3} \exp \left(-\frac{1}{2}\left\|x-x_{n}\right\| \sigma_{s}^{-2}\right)
\end{aligned}
\label{eq:gaussian spare basis}
\end{equation}

where $\omega_{n}$ is the activated value sampled from Gaussian distribution $\mathcal{N}(0, 1)$; $\mu_{n}$ is the basis function centered at a small set of spatially distributed sources, corresponding to presupposition 2, $\boldsymbol{x_n},n=1,...,N_{p}$; $\sigma_s$ is the spatial standard deviation.

The simulated EEG sources are imported to a head model to generate scalp EEG data. Here we segmented an MR template image into 12 tissue classes (skin, eyes, muscle, fat, spongy bone, compact bone, cortical/subcortical gray matter, cerebellar gray matter, cortical/subcortical white matter, cerebellar white matter, cerebrospinal fluid, and brain stem). The head model is solved by the finite element method using the Simbio FEM in the Fieldtrip toolbox \cite{vorwerk2018fieldtrip}. The performance of this head model has been validated by~\cite{liu2018detecting, RN674}. The source dipoles are grid at 6mm by 6mm resolution in the gray matter, with three possible orientations (\ie x, y, z). We further define dipole directions based on the leadfield matrix with a loose hyperparameter $l$. Assuming the principal orientation $d$ of source dipoles can be characterized by leadfield matrix $K$:

\begin{equation}
\begin{aligned}
d =  \sum_{M}K
\end{aligned}
\label{eq:dipole_orientation}
\end{equation}

Here the leadfield matrix $K \in \mathbb{R}^{(M \times 3) \times N}$ where $M$, $N$ represent the number of EEG channels and sources respectively.  The principal orientation $d$ represents the single source orientation towards the scalp sensors.
By randomly sampling unit direction vector $Ori$ and activate value vector $Act$ for sources, we are able to constrain dipole orientation using this practical reference:
\begin{equation}
\begin{aligned}
{\cal M}= 
\left\{\begin{array}{c}
(1) \operatorname{Act}\left[(1-l) d+l \cdot \text { Ori }\right] \text { if } l \cdot \text { Ori }>0 \\
(-{1}) \operatorname{Act}\left[(1-l) d+l \cdot \text { Ori }\right] \text { if } l \cdot \text { Ori } \leq 0
\end{array}\right.
\end{aligned}
\label{eq:loose_orientation}
\end{equation}

After obtaining the leadfield matrix and the corresponding source space, we can generate the scalp EEG according to Eq.\ref{eq:forward}. We randomly sample 1 to 5 dipole sources as the activation centers and generate single activation Gaussian sparse bases $\mu_1, \mu_2\ \text{to}\ \mu_k$. To be more realistic, the addictive Gaussian white noise is added at both source and sensor levels, resembling the background neural activities and the measurement noise respectively. The SNR is set based on the power at sensor level to be 5dB, 10dB and 20dB ~\cite{RN2}:
\begin{equation}
\begin{aligned}
SNR = 10 \log \left(\frac{R M S_{\text {signal }}^{2}}{R M S_{\text {noise }}^{2}}\right)
\end{aligned}
\label{eq:Signal noise ratio}
\end{equation}
The synthetic signals can be finally represented by:
\begin{equation}
\begin{aligned}
\hat{\Phi}=\mathcal{K}\left(\mathcal{M} \Omega+n_{\text {source }}\right)+n_{\text {channel }}
\end{aligned}
\label{eq:synthetic}
\end{equation}
where $n_{\text {source }}$,$n_{\text {channel }}$ represent the Gaussian white noise. 

In this study, we use a three-layer MLP with Rectified Linear Unit (ReLU) as the feature extractor. Dropout and weight decay are used to improve network generalization ability. Weight decay is not used in the parameters of the basis function to avoid the conflict of the edge sparse loss. 

The neural networks are trained in synthetic dataset (loose = 0.1, SNR = 5). The test data is the rest 10\% of synthetic data which are not used for training. We test the ESBN Supervised, ESBN Unsupervised, MNE, dSPM, sLORETA, and eLORETA methods. We also try iteratively reweighted edge sparsity minimization(IRES) method with adapted matrix $V$. As it was developed based on the BEM head model, the IRES result is not ideal under our FEM head model.

\subsection{Experiments on real EEG data}\label{exp:real}

We further validate the performance of our network on real 128-channel EEG data at a motor task. The data was recorded for another study ~\cite{RN918}. The ethics were approved by KU Leuven. We compare our model with other commonly used non-parametric methods (\ie MNE~\cite{RN927}, dSPM~\cite{RN925}, sLORETA~\cite{RN926}, eLORETA~\cite{RN924}). These results are implemented using MNE-Python and Nilearn packages~\cite{RN923,abraham2014machine}. 

The raw EEG data is preprocessed by a standard workflow for hdEEG analysis, including 1-50Hz band-pass filtering and ICA denoising, to obtain the clean EEG signals. The noise covariance is estimated based on 200-250Hz band-pass filtering with the raw EEG signal\cite{RN674}. We then generate a 12-layer volumetric head model using FEM, similar to the method used in synthetic data. 


Since the size of real data is limited and deep learning requires a large dataset for training, we pre-train ESBN in a supervised manner on the simulated dataset, and then use real data to fine tune the parameters of ESBN with the loss function for unsupervised learning (Eq.\ref{eq:loss-unsupervised}).

We randomly sample 100 time points during the motor task as the test data. Based on the clean EEG, the noise covariance and the head model, we estimate the EEG source localizations using our proposed deep learning method, as well as other numerical methods (MNE, dSPM, sLORETA, eLORETA). 

\subsection{Metrics to quantify performance}
\label{sub:metrics}
We use the localization error, spatial dispersion, area under curve as metrics to quantify EEG source localization performance.

The localization error (LE) can be quantified as the Euclidean distance between truly activated source $r_{true}$ and the reconstructed peak source $r_{peak}$ in three dimensional source space:
\begin{equation}
\begin{aligned}
\begin{array}{l}
\mathrm{LE}=\left\|\mathbf{r}_{true}-\mathbf{r}_{peak}\right\|_{2} \\
\end{array}
\end{aligned}
\label{eq:localization error}
\end{equation}

The spatial dispersion (SD) is another metric for EEG source localization. SD can be represented by~\cite{samuelsson2021spatial}:
\begin{equation}
\begin{aligned}
\mathrm{SD}=\frac{\sum_{k=1}^{N} d_{j k}\left|\hat{\mathbf{x}}_{k}\right|}{\sum_{k=1}^{N}\left|\hat{\mathbf{x}}_{k}\right|}, d_{j k}=\left\|\mathbf{r}_{j}-\mathbf{r}_{k}\right\|_{2}
\end{aligned}
\label{eq:spatital dispersion}
\end{equation}
where $d_{j k}$ is a result matrix of Eq. \ref{eq:localization error}, reflecting  the localization error (LE) value for all sources. Notably, a lower spatial dispersion reflects a better ability to locate multiple sources~\cite{samuelsson2021spatial}. 

The area under curve (AUC) calculates the area under the precision-recall characteristics curve, which is composed of true positive rate and false positive rate, makes the value between 0 (always wrong) and 1 (always right)~\cite{samuelsson2021spatial}. 

For the tests on synthetic data, the source localization performance is quantified using LE, SD and AUC. For the tests on real EEG data during the motor task, the performance is quantified by source spatial dispersion, since no ground truth is provided. 

\section{Results}
\label{S:results}
\subsection{Results on simulated data}
Table \ref{tab1} shows the quantitative results from the synthetic dataset. The results suggest that the ESBN Supervised method outperforms other methods. It implies that the network structure of ESBN has a considerable representation ability for EEG source imaging. Although the ESBN Unsupervised is as good as the numerical algorithm, they are on a similar scale. In the following tests, the performance of the model under different experimental conditions will be examined.

\begin{figure*}%
\centering
\subfigure[Localization error for sources with different depths]{
\includegraphics[width=0.8\columnwidth]{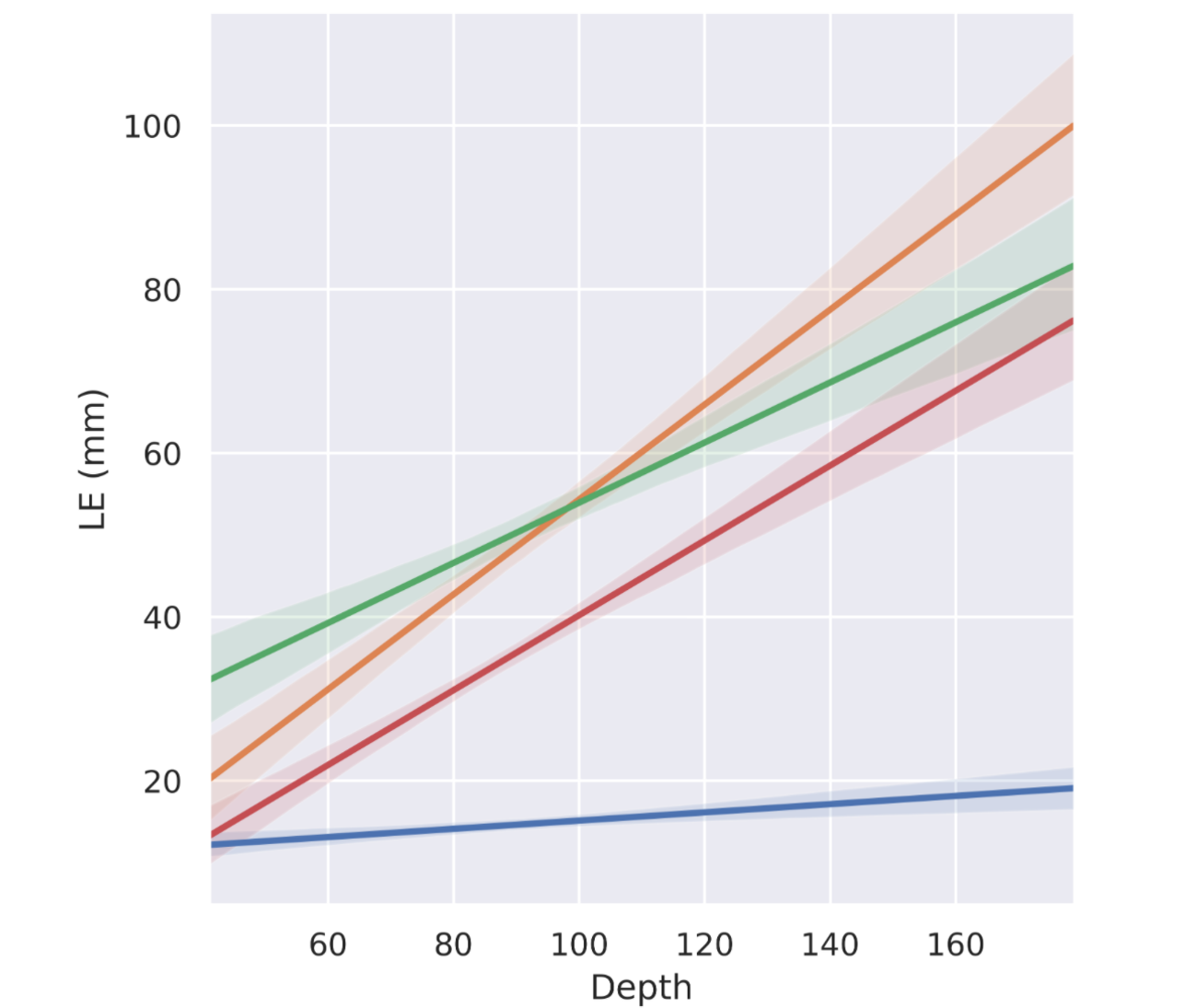}}
\subfigure[AUC for sources with different depths]{
\includegraphics[width=0.8\columnwidth]{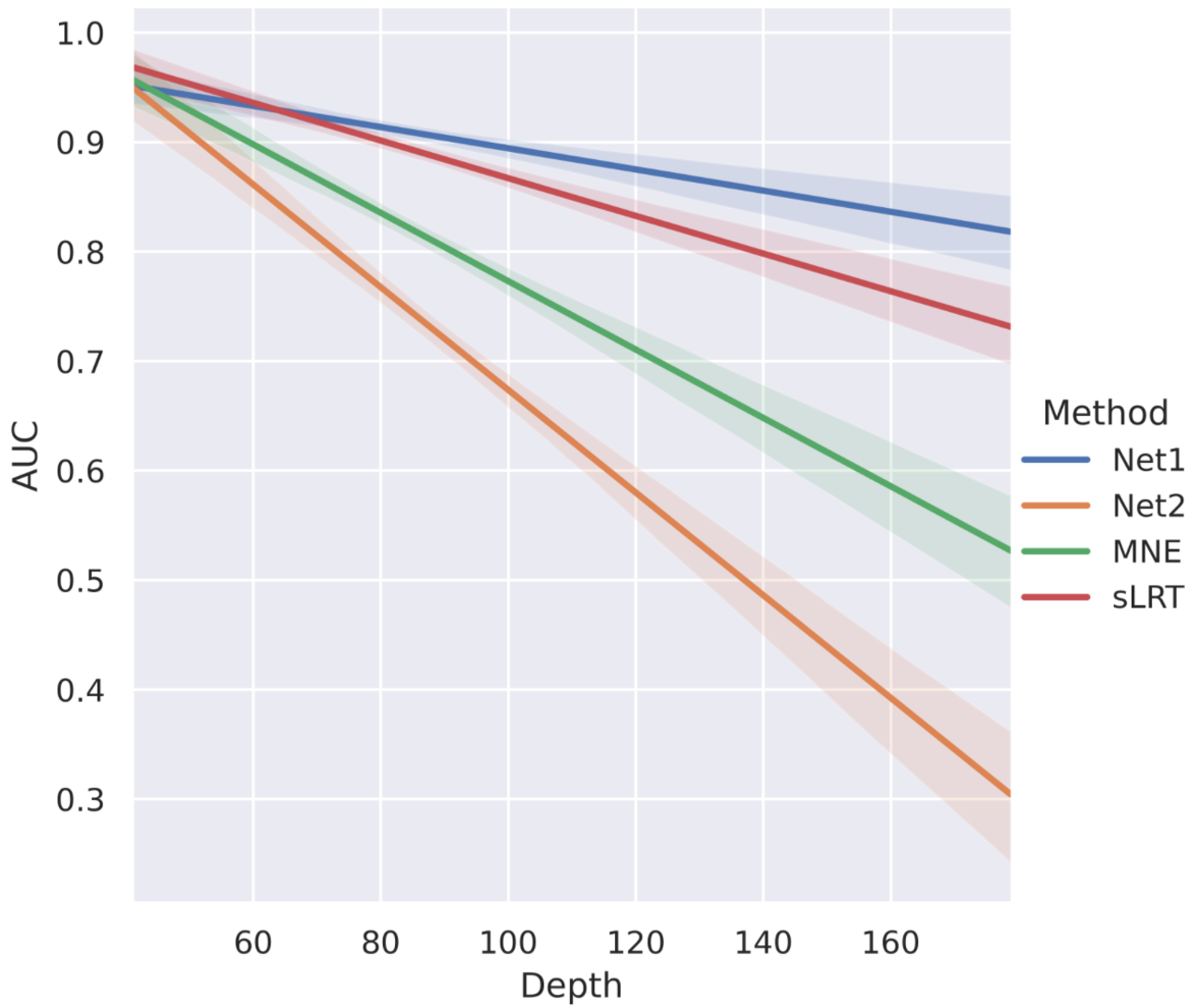}}
\caption{The impacts of source depth to the localization accuracy. \textit{Abbrev.}: Net1, ESBN Supervised; Net2, ESBN Unsupervised; sLRT, sLORETA.}
\label{fig.depth}
\end{figure*}

The influence of the depth of simulated EEG source on the source localization performance is tested, and the result is shown in Fig.\ref{fig.depth}. To be noted, the source depth is defined by the sum of the corresponding column in the leadfield matrix, rather than the anatomical depth. The larger the depth is, the weaker the source affects the scalp signal. The supervised ESBN (Net1) has a great ability to reconstruct the deep sources, while the performance of other methods (unsupervised ESBN, MNE and eLORETA) is largely reduced with increasing depth.

\begin{figure*}%
\centering
\subfigure[LE for signals with different SNRs]{
\includegraphics[width=0.8\columnwidth]{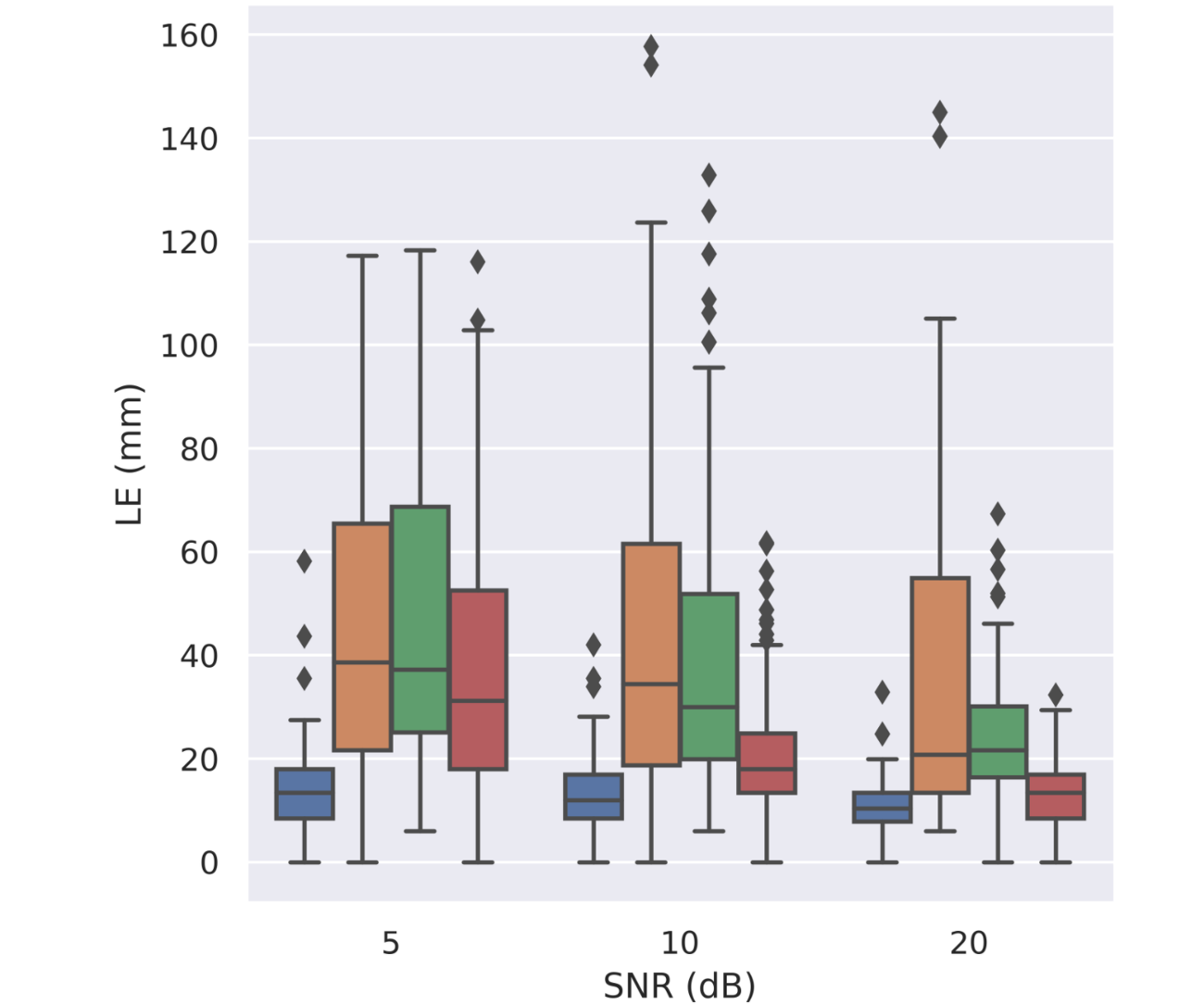}}
\subfigure[AUC for signals with different SNRs]{
\includegraphics[width=0.8\columnwidth]{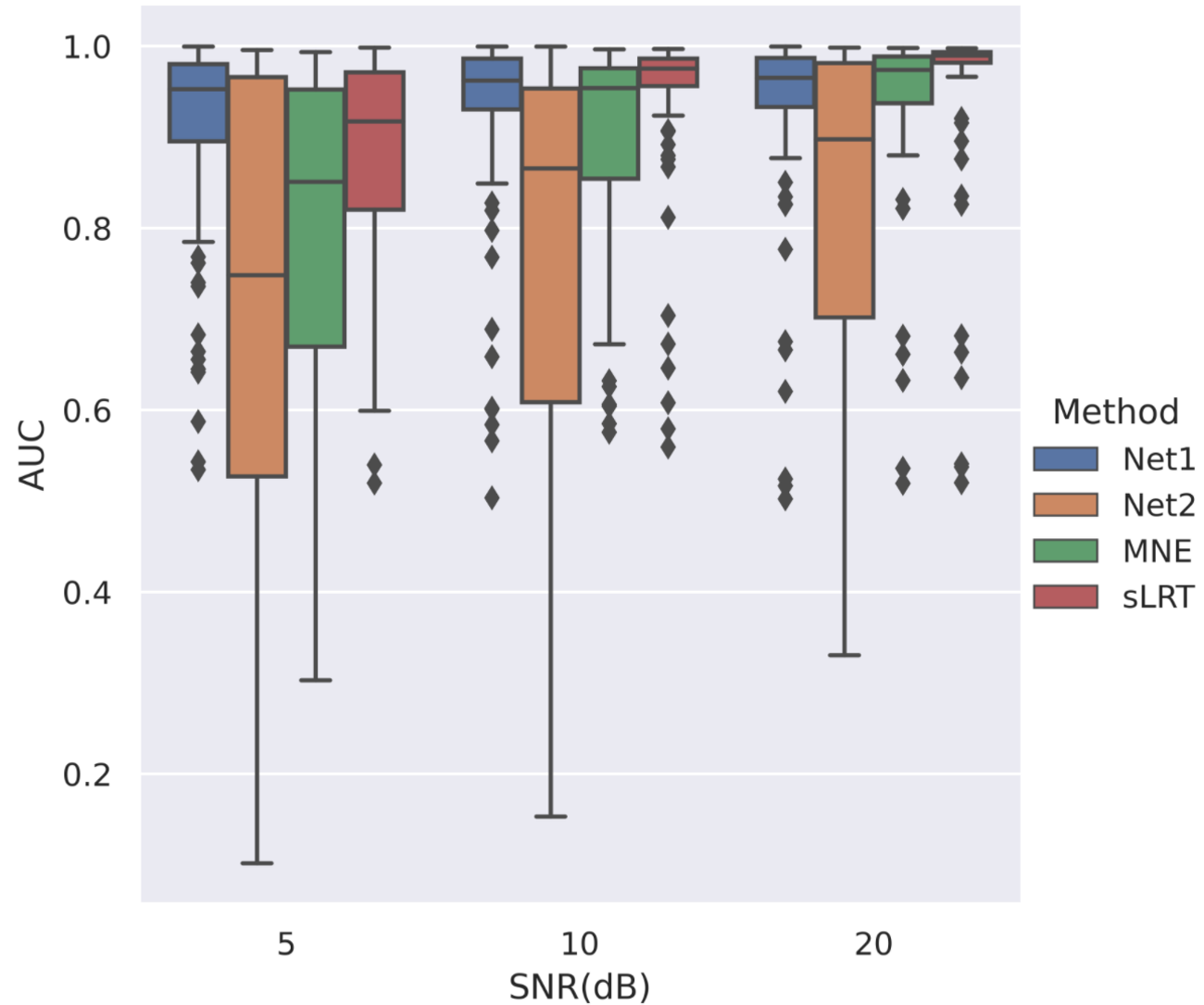}}
\caption{The effect of SNR on localization performance. We vary the signal-to-noise ratio $SNR$ at sensor level (Eq.\ref{eq:Signal noise ratio}). }
\label{fig.SNR}
\end{figure*}

Fig. \ref{fig.SNR} shows the impacts of noise on EEG source localization. Compared with numerical algorithms, our model is more robust to noise, as it adopts many maneuvers from the deep learning community to cancel the effects of noise, such as dropout and weight decay. 

\begin{figure*}%
\centering
\subfigure[LE for signals generated with different loose factors]{
\includegraphics[width=0.8\columnwidth]{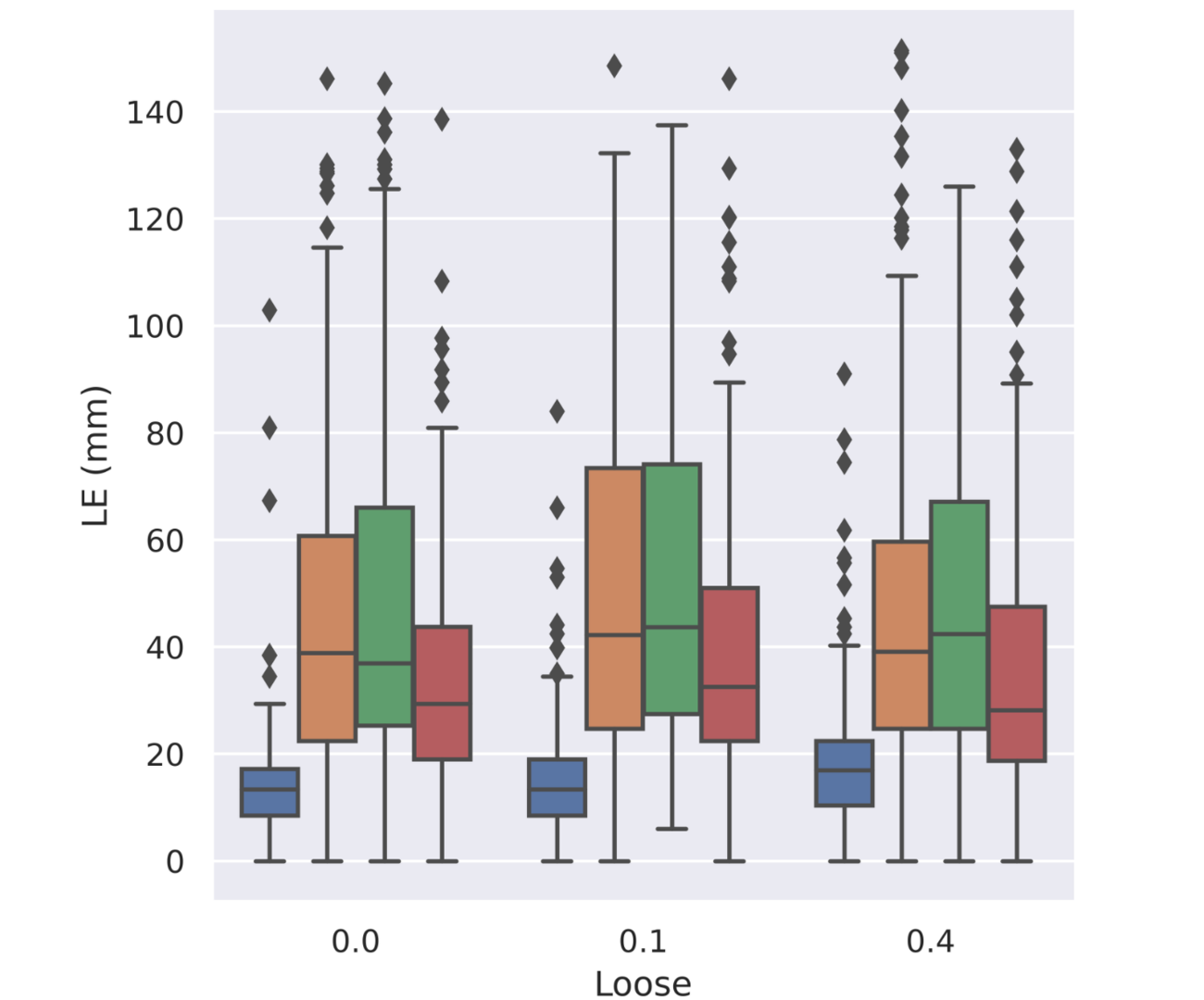}}
\subfigure[AUC for signals generated with different loose factors]{
\includegraphics[width=0.8
\columnwidth]{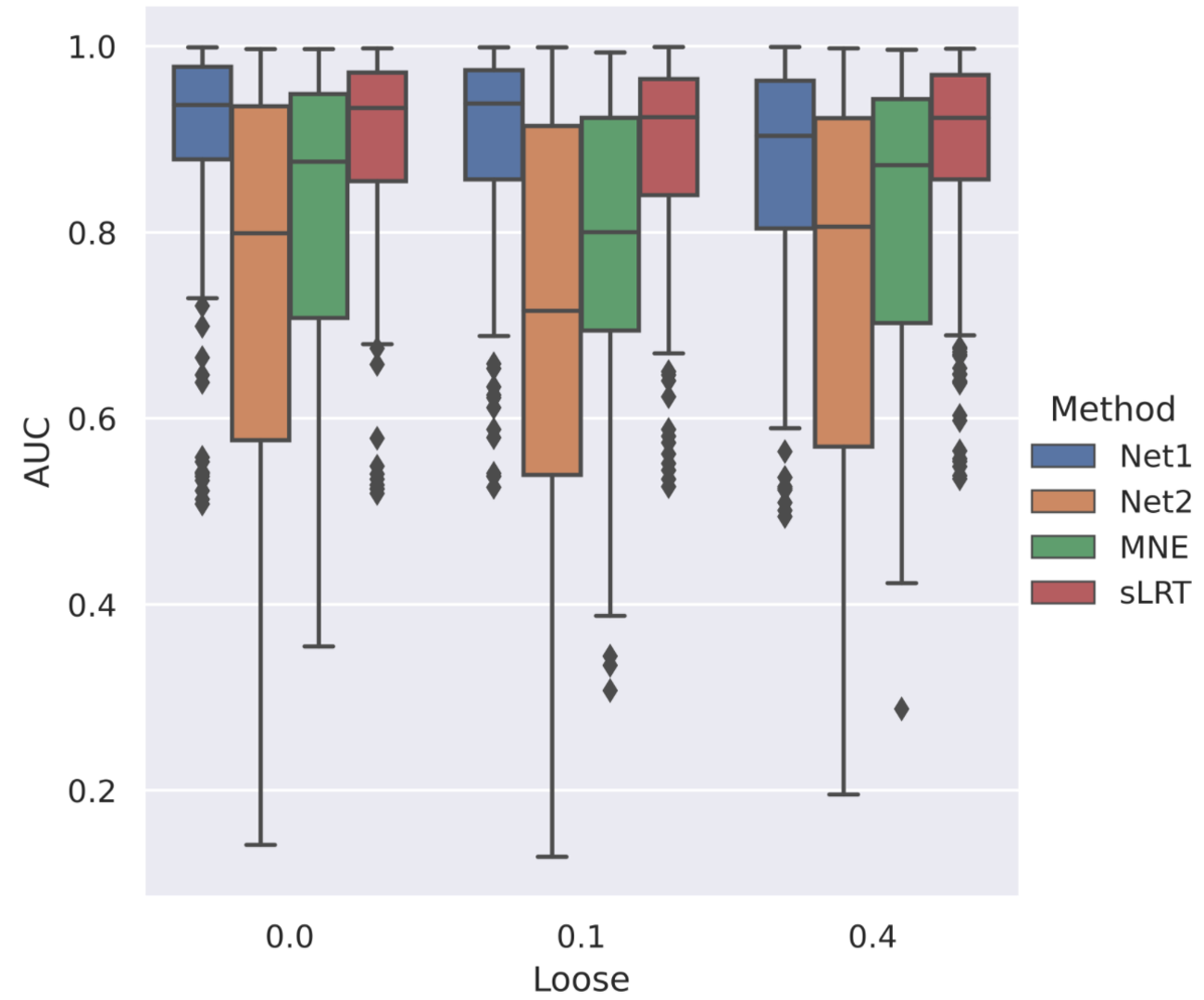}}
\caption{The effect of dipole direction settings on localization performance. We vary the loose factor $l$ (Eq.\ref{eq:loose_orientation}) for EEG simulations. }
\label{fig.LOOSE}
\end{figure*}

Since the data simulation process relies on the constraint on the dipole orientations, we test their effects on the source localization by varying the loose factor $l$ during simulation. Fig. \ref{fig.LOOSE} shows the influence of dipole orientation. The Supervised ESBN is most affected, as the loose factor directly changes the distribution of the source space and the end-to-end learning is sensitive to the target space distribution. In contrast, the loose factor does not have a visible effect on the other methods. 

\begin{table}[]
\caption{Evaluation of Methods on Synthetic Data }
\begin{center}
\begin{tabular}{|l|c|c|c|}
\hline
\textbf{Methods} & \textbf{LE}& \textbf{SD}& \textbf{AUC} \\
\hline
\textbf{ESBN Supervised $^*$}& \textbf{14.98(10.63)}& \textbf{21.96(8.02)} & \textbf{0.91(0.11)} \\
\hline
\textbf{ESBN Unsupervised}& 46.83(32.51)& 75.54(7.4) & 0.72(0.23) \\
\hline
\textbf{MNE}& 49.04(31.36) & 64.34(13.82)& 0.81(0.17) \\
\hline
\textbf{dSPM}&35.42(12.98) &48.48(7.87) & 0.88(0.11) \\
\hline
\textbf{sLORETA} &34.84(23.95) &66.39(11.44) & 0.89(0.11) \\
\hline
\textbf{eLORETA} &38.58(26.35) &67.37(11.25) & 0.88(0.12) \\
\hline
\end{tabular}
\end{center}
\tiny $^*$ indicates the winner method. The numbers in the brackets are the standard deviation. \textit{Abbrev.}:LE, Localization Error; SD, Spatial Dispersion; AUC, Area Under Curve. (Simulation dataset parameters: Loose $l$ = 0.1, SNR = 5)
\label{tab1}
\end{table}

\subsection{Results on real EEG data}
\begin{figure*}%
\centering
\includegraphics[width=0.78\textwidth]{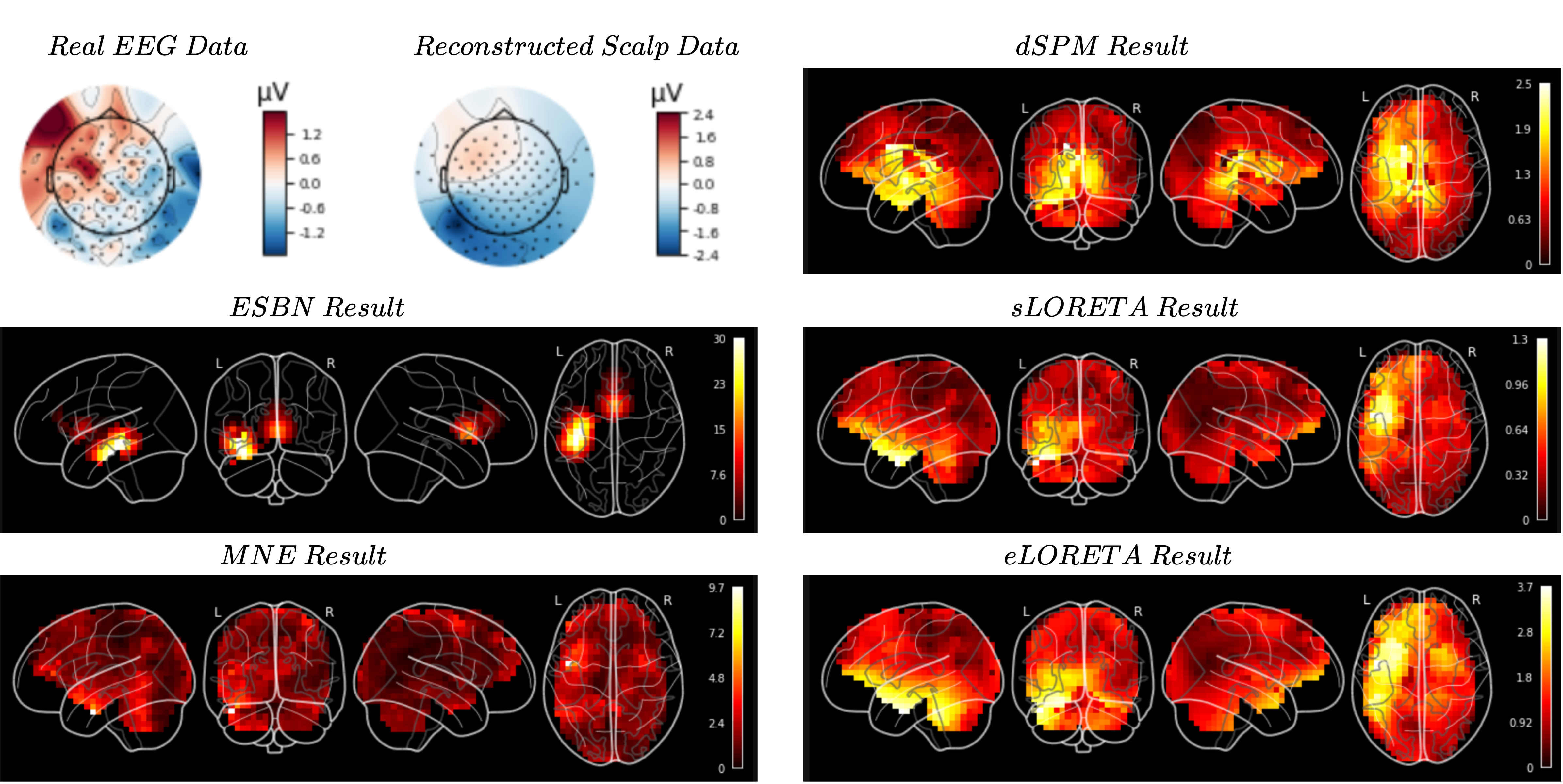}
\caption{Performance on real EEG data. (top left) the real EEG topomap and the EEG topomap generated through the forward model with the estimated sources by ESBN; (other panels) the localized EEG sources based on ESBN, MNE, dSPM, sLORETA and eLORETA methods.}
\label{fig.realdata}
\end{figure*}

Based on the preprocessed clean EEG signal, We compare ESBN performance with the traditional numerical methods (\ie MNE, dSPM, sLORETA and eLORETA). An example of localized sources is presented in Fig.~\ref{fig.realdata}. It demonstrates that compared to numerical results, the ESBN localized sources are more sparse and focused on the motor regions. It also implies that the EEG sources can be recognized as a combination of independent basis functions. In return, the reconstructed scalp EEG (Fig.\ref{fig.realdata} top left) that based on ESBN inferred sources and 12-layer FEM forward model has a smoother topological distribution, possibly due to the elimination of channel noise. As we lack ground truth for real EEG sources, we use the spatial dispersion of source localization to quantify the performance, shown in Table ~\ref{tab2}.

\begin{table}[]
\caption{Evaluation of Methods on Real Data}
\begin{center}
\begin{tabular}{|l|c|c|c|c|c|}
\hline
\textbf{SD} & \textbf{ESBN}& \textbf{MNE}& \textbf{dSPM}& \textbf{sLORETA}& \textbf{eLORETA} \\
\hline
\textbf{Value}& \textbf{43.76}& 82.86 &59.22& 71.61&77.01 \\
\hline
\textbf{Std}& 13.41& 5.57 &8.87& 7.54&7.65 \\
\hline
\end{tabular}
\end{center}
\label{tab2}
\end{table}

\section{Discussion and future directions}
\label{S:discussion}
In this paper, we propose a deep learning framework called ESBN for EEG source localization. Although our framework is based on edge sparsity prior and Gaussian basis function, it can be easily extended to other priors by adjusting the term ${\cal S}()$ in Eq.~\ref{eq:loss-reg}. 
The network can be trained on a large synthetic dataset in a supervised way, and it can be further refined using a small real EEG dataset in an unsupervised way for better generalization. 

Performance of ESBN is validated using both synthetic data and real EEG data. The results in synthetic data suggest that the end-to-end supervised ESBN outperforms the network trained in an unsupervised way and other non-parametric methods (Table~\ref{tab1}), in terms of reconstructing deep sources (Fig.~\ref{fig.depth}) and robustness to noise (Fig.~\ref{fig.SNR}). This is along with our previous study in machine learning~\cite{yin2021riemannian}. It has been well documented that the deeper sources are more difficult to reconstruct ~\cite{grech2008review} for both numerical algorithms and neural networks. Importantly, our method shows a strong capability for deep sources (Fig.~\ref{fig.depth}), as the end-to-end supervised ESBN can directly learn the prior distribution of the sources. Although lacking ground truth in real datasets would weaken the learning ability, we can use a transfer learning design to adjust the prior distribution for adaptation to real data. The unsupervised training using the loss function defined on the sensor level can help improve generalizability to real EEG data (Fig.~\ref{fig.realdata}). The unsupervised training procedure can reshape network parameters based on real data, which might leverage the source localization for abnormal EEG data whose sources might not be simulated in synthetic data, such as epilepsy.  

Despite the advantages of ESBN, it is worthy to note the limits. First, the dipole orientations in the volumetric head model need to be constrained, as the free orientation can greatly influence the source imaging results (Fig.~\ref{fig.LOOSE}). Here we use a loose factor on PCA to constrain the dipole orientation, but other alternative ways worth to be further investigated. Second, we only consider the stationary spatial information in ESBN, and do not incorporate temporal information of source dynamics. One way is to add regularization terms based on the stability of source time series~\cite{gramfort2013time}; however, this might come at a cost of generalizability. Another way to employ recurrent neural network (RNN) to extract temporal information. This will be our future direction. 

The ill-posed nature of EEG source localization is rooted in our poor understanding of the internal neurophysiological mechanisms. With a better understanding of neural source dynamics and distribution, we can design more realistic priors. On the other hand, advanced neural recording techniques can be beneficial as it provides rich information. For instance, simultaneous EEG-fMRI can facilitate EEG source localization with fMRI signal acting as high spatial resolution priors to constrain the source space~\cite{lei2011fmri}. The invasive neural recordings (such as ECoG or SEEG) might provide ground-truth source dynamics for EEG source localization. 

\section*{Acknowledgment}
The authors declare no competing interests.

\bibliographystyle{IEEEtran}
\bibliography{IEEEabrv, Myreference}

\end{document}